\pdfoutput=1

\documentclass[11pt]{article}

\usepackage[preprint]{acl}

\usepackage{times}
\usepackage{latexsym}
\usepackage{inconsolata}
\usepackage{tabularx}
\usepackage{float}
\usepackage{lipsum}
\usepackage{amsmath}
\usepackage{amssymb}
\usepackage{multirow}
\usepackage{makecell}
\usepackage{algorithm}
\usepackage{algpseudocode}
\usepackage[dvipsnames]{xcolor}

\usepackage[T1]{fontenc}

\usepackage[utf8]{inputenc}

\usepackage{microtype}

\usepackage{inconsolata}
\usepackage{subcaption}

\usepackage{graphicx}
\usepackage{booktabs}

\usepackage{pifont}      
\usepackage{enumitem}    

\newcommand{\cmark}{\ding{51}}  
\newcommand{\xmark}{\ding{55}}  

\newcommand\ours{\textcolor{black}{\textsc{Entropy2Vec}}}

\title{\ours{}: Crosslingual Language Modeling Entropy \\ as End-to-End Learnable Language Representations}

\author{
Patrick Amadeus Irawan$^{1*}$,
Ryandito Diandaru$^{1*}$,
Belati Jagad$^{2*}$,
Randy Zakya Suchrady$^{3*}$ \\
\textbf{Alham Fikri Aji$^1$,
Genta Indra Winata$^4$,
Fajri Koto$^1$,
Samuel Cahyawijaya$^{5*}$} \\
$^1$MBZUAI \quad
$^2$Universitas Indonesia \quad
$^3$NTU \quad
$^4$Capital One \quad
$^5$Cohere
\\
\texttt{\{patrick.irawan, ryandito.diandaru\}@mbzuai.ac.ae} \\
\small $^*$Main authors
}

\begin{document}
\maketitle

\begin{abstract}

We introduce \ours{}, a novel framework for deriving cross-lingual language representations by leveraging the entropy of monolingual language models. Unlike traditional typological inventories that suffer from feature sparsity and static snapshots, \ours{} uses the inherent uncertainty in language models to capture typological relationships between languages. By training a language model on a single language, we hypothesize that the entropy of its predictions reflects its structural similarity to other languages: Low entropy indicates high similarity, while high entropy suggests greater divergence. This approach yields dense, non-sparse language embeddings that are adaptable to different timeframes and free from missing values. Empirical evaluations demonstrate that \ours{} embeddings align with established typological categories and achieved competitive performance in downstream multilingual NLP tasks, such as those addressed by the LinguAlchemy framework.

\end{abstract}


\section{Introduction}

Linguistic typology provides a framework for classifying languages based on shared structural features, offering insights into language universals and diversity. Databases like the World Atlas of Language Structures (WALS)~\cite{haspelmath2005world}, AUTOTYP~\cite{bickel2002autotypologizing}, URIEL~\cite{littell-etal-2017-uriel}, and URIEL$^+$~\cite{khan-etal-2025-uriel} catalog these features, serving as valuable resources for researchers and practitioners in the field of computational linguistics and  beyond. However, these inventories face significant limitations: they often cover only a subset of languages, leading to missing values, and they represent static snapshots of linguistic knowledge, neglecting the dynamic and evolutionary nature of languages.

\begin{figure}[!ht]
    \centering
    \includegraphics[width=\columnwidth]{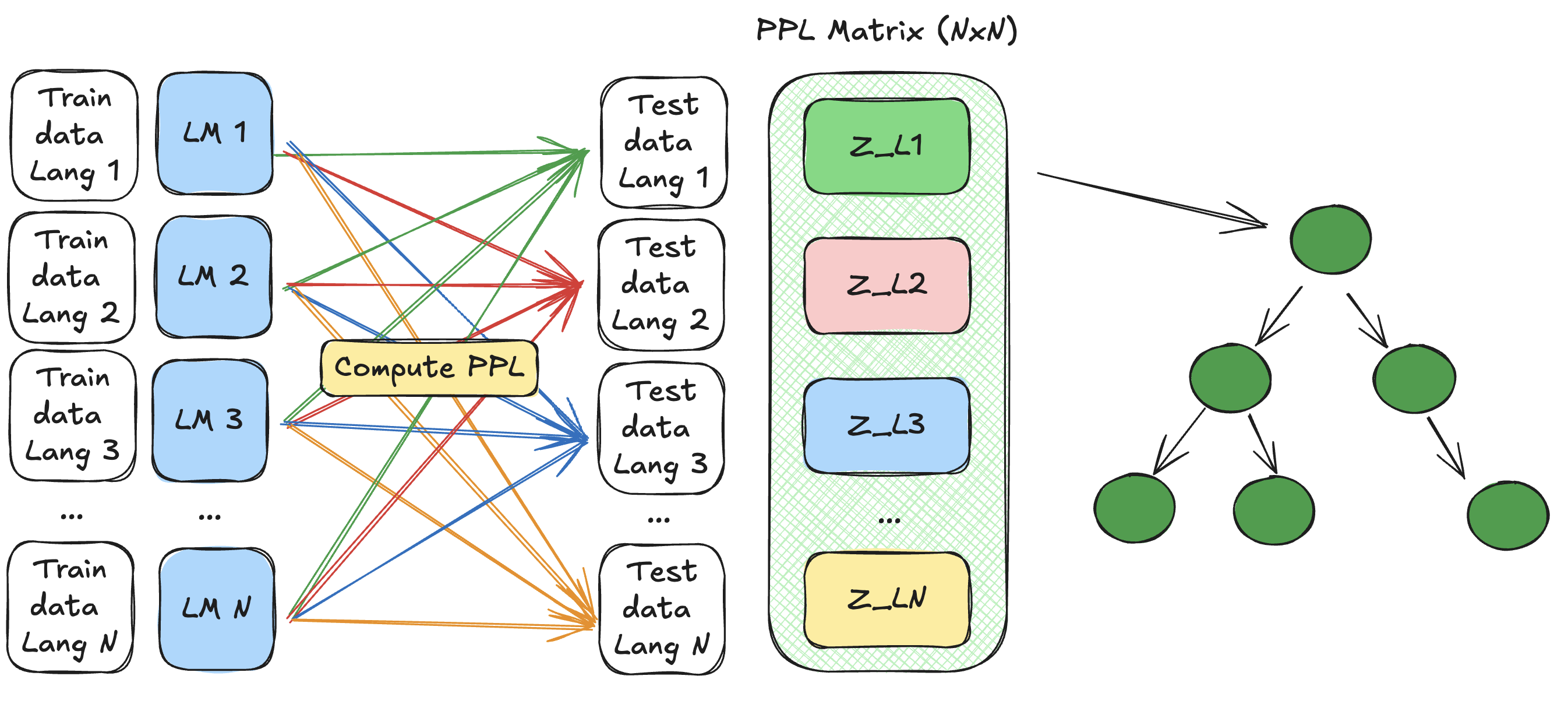}
\caption{\ours{} framework. Monolingual LMs are trained per language, and cross-lingual perplexity is used as an unsupervised signal to derive language vectors and induce typological trees, aligning well with expert-curated taxonomies.}
    \label{fig:fig-intro}
\end{figure}

Recent advancements in neural language modeling have enabled the extraction of continuous representations of languages through pre-trained models. These embeddings capture semantic and syntactic properties, facilitating cross-lingual transfer in various NLP tasks. Nonetheless, existing methods primarily focus on monolingual or bilingual settings and do not explicitly model the typological relationships between languages. Moreover, they often rely on manually curated features, which may not generalize well across languages or over time.

To address these challenges, we propose \ours{}, a framework that derives language representations based on the entropy of monolingual language models (LMs). Entropy, a measure of uncertainty in information theory, reflects the predictability of a language's structure. By training a language model on a single language and analyzing its entropy when applied to other languages, we can infer typological similarities and differences. This approach offers several advantages: it is data-driven, scalable, and inherently adaptable to new languages and evolving linguistic features.

In this paper, we demonstrate that \ours{} embeddings align with established typological categories, such as phonological, morphological, and syntactic features. We also show that these embeddings outperform traditional typological inventories in downstream multilingual NLP tasks, including language identification, typology prediction, and cross-lingual transfer. By integrating \ours{} into the LinguAlchemy framework~\cite{adilazuarda2024lingualchemy}, we achieve competitive generalization across languages, especially those underrepresented in existing typological resources.

\section{Related Works}
\paragraph{Typological Language Inventories} Traditional typological inventories, such as WALS~\cite{haspelmath2005world}, AUTOTYP~\cite{bickel2002autotypologizing}, URIEL~\cite{littell-etal-2017-uriel}, and URIEL$^+$~\cite{khan-etal-2025-uriel}, have been instrumental in documenting linguistic diversity and informing computational models. However, these resources are limited by their static nature and the incomplete coverage of the world's languages. For instance, WALS provides typological data for only a fraction of the estimated 7,000 languages, leading to missing values that can hinder the performance of NLP models . \ours{} addresses these limitations by deriving LMs from the entropy of monolingual LMs. This approach is inherently dynamic, as it can adapt to new languages and evolving linguistic features without the need for manual curation. Moreover, it provides dense, non-sparse embeddings that capture the probabilistic structure of languages, offering a more nuanced understanding of typological relationships.

\paragraph{Language Vector in NLP} 

Language vectors, or embeddings, have become foundational in modern NLP, enabling models to represent words, sentences, and even entire languages as continuous vectors in a high-dimensional space. Techniques like Word2Vec, GloVe, and FastText have demonstrated that such embeddings capture semantic and syntactic properties, facilitating tasks like word similarity, analogy reasoning, and machine translation. These embeddings are typically learned from large corpora and reflect the statistical patterns in language use. However, they often treat languages as isolated entities, without explicitly modeling the relationships between them. Recent advancements, such as multilingual BERT and XLM-R, have sought to address this by training models on multiple languages simultaneously, capturing shared structures and enabling cross-lingual transfer.\ours{} contributes to this landscape by offering a novel perspective on language representation. Instead of relying solely on large-scale pre-training on vast corpora, \ours{} leverages the entropy of monolingual LMs to infer typological relationships between languages. This approach not only aligns with existing language representation models but also extends their capabilities by incorporating typological insights, thereby enhancing multilingual understanding and transfer learning

\section{\ours{}}

\subsection{Unsupervised Language Modeling}

Unsupervised language modeling uses an autoregressive approach, where the LM predicts the next token based on the previous ones. Mathematically, given a sequence of tokens $[x_1, x_2, \dots, x_t]$, the LM defines a probability distribution over the next token $x_{t+1}$ conditioned on all previous tokens. This can be formally expressed as:
\[
    x_{t+1} = \arg\max_{x} P(x \mid x_1, x_2, \dots, x_t; \theta)
\]

where and $\theta$ represents the parameters of the model. The goal of training is to maximize the likelihood of the observed data, which is equivalent to minimizing the cross-entropy loss. Formally, given a dataset $\mathcal{D} = {(x_1^{(1)}, \dots, x_{n_1}^{(1)}), \dots, (x_1^{(N)}, \dots, x_{n_N}^{(N)})}$, the cross-entropy loss is defined as:
\[
\mathcal{L}(\theta, \mathcal{D}) = -\frac{1}{N} \sum_{i=1}^N \sum_{t=1}^{n_i} \log P(x_t^{(i)} \mid x_1^{(i)}, \dots, x_{t-1}^{(i)}; \theta)
\]

This encourages the model $\theta$ to assign high probability to the actual next tokens in the training data. The autoregressive nature of these models allows them to generate coherent and contextually relevant text by sequentially predicting tokens~\cite{radford2019language,brown2020languagemodelsfewshotlearners,cahyawijaya-etal-2021-indonlg}, making them highly effective for building strong language representations~\cite{workshop2023bloom,cohere2025commanda}.

\subsection{LM Entropy as Language Vectors}

Although having a strong language representation, LMs can only produce meaningful representation on languages that they have been pre-trained on~\cite{winata-etal-2023-nusax, cahyawijaya-etal-2023-instructalign} ans closely similar languages~\cite{cahyawijaya-etal-2023-nusawrites,cahyawijaya-etal-2024-cendol}. The cross-lingual generalization often diminish when the corresponding model is faced with languages that are low-resource~\cite{bang-etal-2023-multitask,cahyawijaya-etal-2023-nusacrowd} and distant from the languages it has been trained on~\cite{lovenia-etal-2024-seacrowd,cahyawijaya2024llmeveryone,bean2024lingoly}. 

As the cross-lingual generalization of LMs depends on the closeness of the language, we argue that this limitation can actually be exploited to build a language vector which is a vector that provides a global representation of a certain language. More specifically, using a set of monolingual LMs $\{ \theta_{L_1}, \theta_{L_2}, \dots, \theta_{L_n}\}$ each trained on a specific language $L_i$ and a set of monolingual corpora $\{ \mathcal{D}^{L_1}, \mathcal{D}^{L_2}, \dots, \mathcal{D}^{L_n} \}$, we build the vector representation of languages $\{Z^{L_1}, Z^{L_2}, \dots, Z^{L_n}\}, Z^{L_i} \in \mathbb{R}^{n}$ by computing the average cross-entropy of the corresponding language model $\theta_i$ on each corpus $\mathcal{D}_j$. Formally, we define the language vector $Z^{L_i}$ as:
\[
Z^{L_i} = \left[ \mathcal{L}(\theta_i, \mathcal{D}_1),  \mathcal{L}(\theta_i, \mathcal{D}_2), \dots,  \mathcal{L}(\theta_i, \mathcal{D}_n) \right]
\]

We call our method of deriving language vector from the entropy of LMs as \ours{}. Unlike other existing language vectors like URIEL~\cite{littell-etal-2017-uriel} and URIEL$^+$~\cite{khan-etal-2025-uriel}, which derive their language vectors from various linguistic inventories, e.g., WALS~\cite{wals}, AUTOTYP~\cite{AUTOTYP}, etc., our method provides a fully unsupervised, data-driven approach for building a language vector. Moreover, our vector can evolve following the actual evolution of languages by updating each of the monolingual LMs with more recent data on each of the corresponding languages. \ours{} leverages the inherent patterns and structures within large-scale textual data, eliminating the need for manual feature engineering or reliance on predefined linguistic inventories. By continuously updating the models with new data, our approach ensures that the language vectors remain dynamic and reflective of the ever-changing nature of human language.

\section{\ours{} and Language Typology}
To assess the validity of \ours{}, we compare it against several established language vector and tree baselines: URIEL~\cite{littell-etal-2017-uriel} and URIEL$^+$~\cite{khan-etal-2025-uriel} vectors, as well as the Glottolog tree ~\cite{nordhoff2011glottolog}. For the first two, we derive a hierarchical clustering tree representing inter-language distances based on geographical and syntactic features. We then evaluate how well the trees induced from \ours{} vectors replicate these known typological groupings, and whether they reveal novel or diverging relationships.

\subsection{Experiment Setting}

\paragraph{Dataset}
Our data source is the Glot500c corpus~\cite{imanigooghari-etal-2023-glot500}, from which we gather textual data for 33 distinct languages which are also present in URIEL, URIEL$^+$, and Glottolog.
For each language, we cap the data at a maximum of 1M sentences and split this data into 7:2:1 (train, validation, test) split after collating the sentence to cap each instance to 1024 characters to support model's max ingestion length. The details of the quantity and split per language can be observed in Appendix \ref{app:dataset-split}.

\paragraph{Training Strategy}
We choose GPT-2 as our pre-trained language model for learning language representations, where the model is configured with an embedding dimension of 512, 4 transformer layers, and 8 attention heads. More details—including tokenizer configurations, optimization parameters, and the precise methodology for perplexity extraction—are elaborated further in Appendix \ref{app:training-hyperparams}. Training is conducted by using the same settings for all 33 languages to extract their perplexity, a measure of how well the language model predicts the test data. This perplexity scores, reflecting the model's "surprise" by a language's characteristics, are used to derive language vectors denoted as $\{Z^{L_1}, Z^{L_2}, \dots, Z^{L_n}\}$, where each $Z^{L_i}$ represents a specific language. From now on, the entirety of these vectors will be termed as \ours{}.

\begin{figure*}[!th]
    \centering

    \begin{subfigure}[t]{0.98\textwidth}
        \centering
        \includegraphics[width=\linewidth]{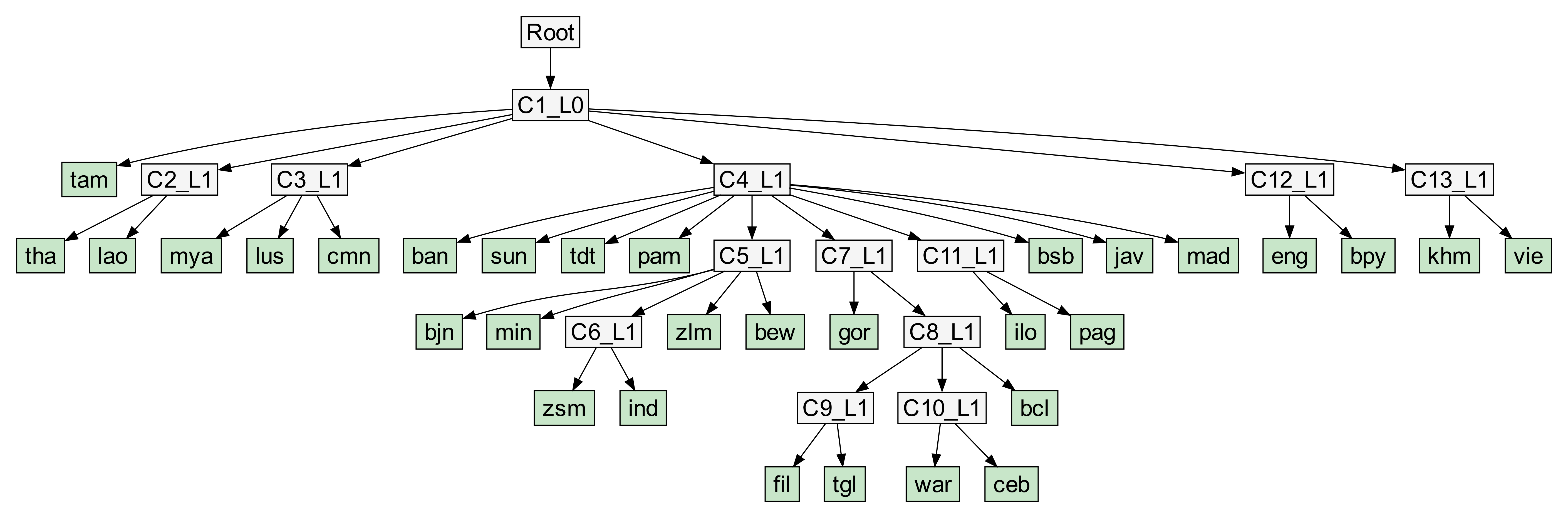}
        \caption{Glottolog Tree (pruned)}
        \label{fig:glottolog-tree}
    \end{subfigure}

    \vspace{1em} 

    \begin{subfigure}[t]{0.49\textwidth}
        \centering
        \includegraphics[width=\linewidth]{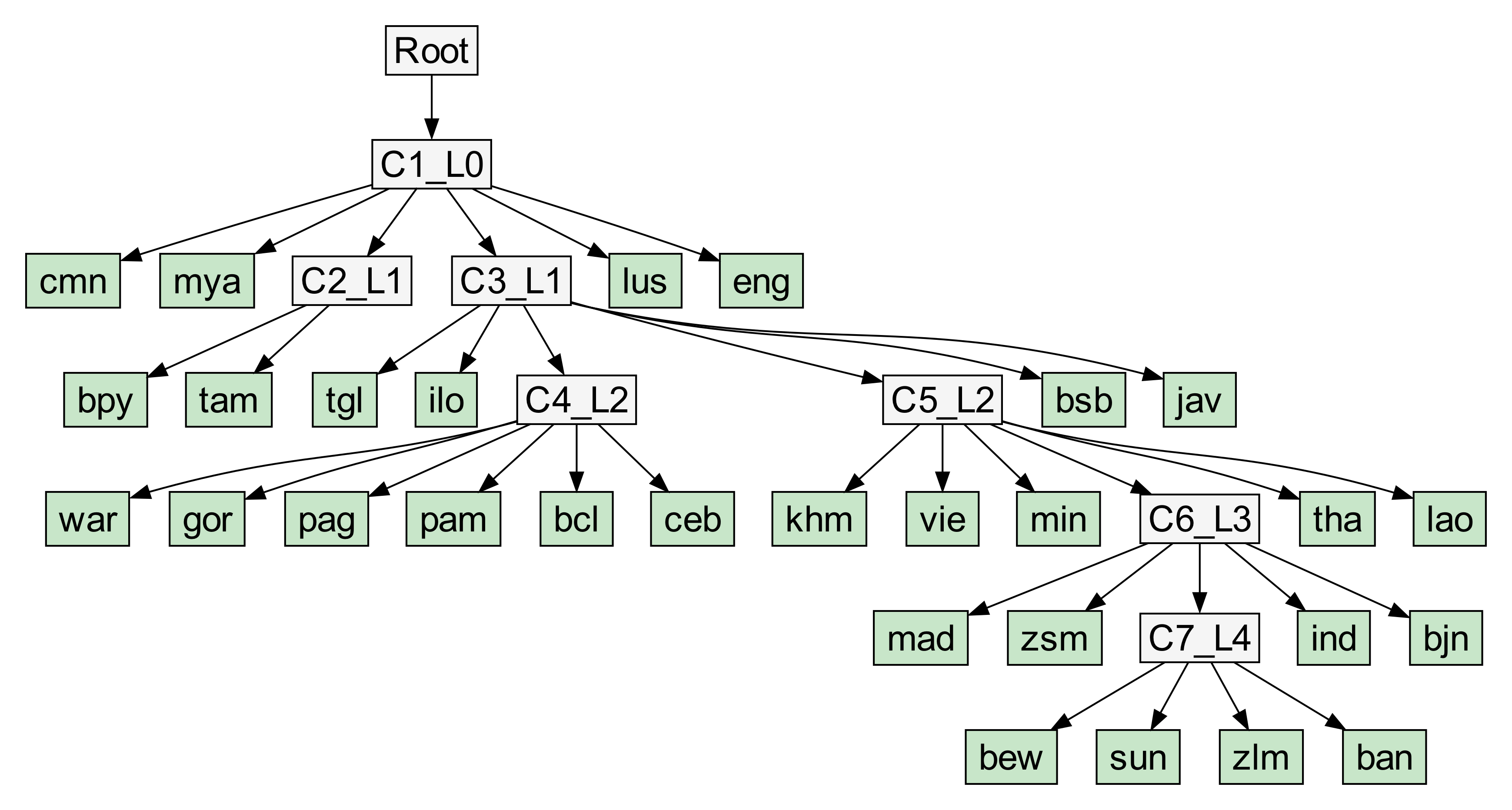}
        \caption{Tree from URIEL \texttt{syntax-knn} (pruned)}
        \label{fig:uriel-tree}
    \end{subfigure}
    \hfill
    \begin{subfigure}[t]{0.49\textwidth}
        \centering
        \includegraphics[width=\linewidth]{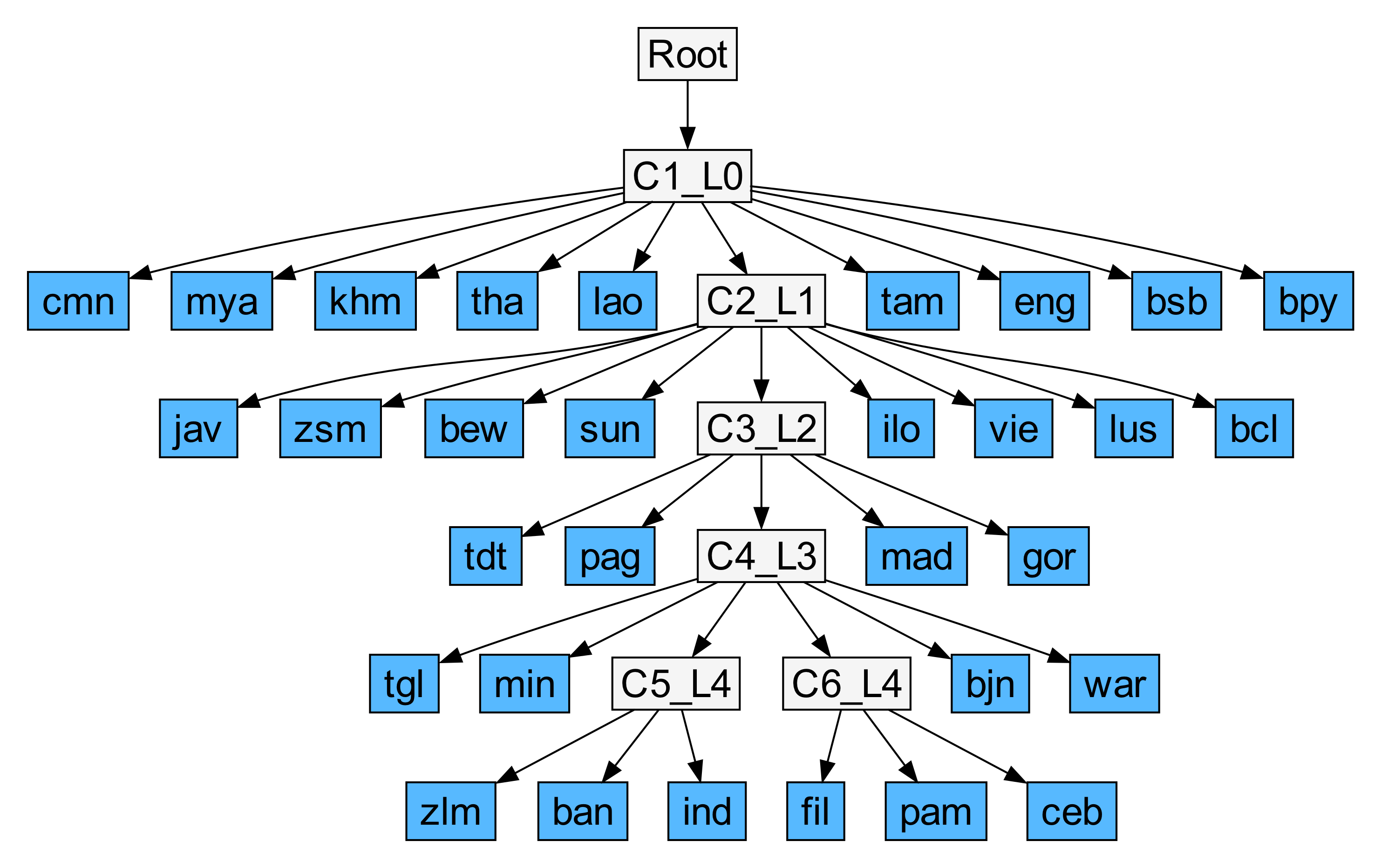}
        \caption{Tree from \ours{}}
        \label{fig:ppl-tree}
    \end{subfigure}

    \caption{Tree comparison across methods: (a) Glottolog gold tree, (b) tree derived from URIEL \texttt{syntax-knn} distances, and (c) our tree derived using \ours{} representations with perplexity-based clustering.}
    \label{fig:tree-all}
\end{figure*}

\paragraph{Forming Typological Trees}  
We generate hierarchical language clusters from the learned vector representations $Z^{L}$using the DBSCAN algorithm, selected for its ability to discover clusters of arbitrary shape without requiring a predefined number of clusters. This choice is motivated by the non-uniform density and structure of real-world language typologies, which traditional linkage-based methods fail to capture due to its complexity (Appendix \ref{app:linkage-failure}). The resulting clusters are then transformed into tree structures and post-processed to ensure compatibility with downstream evaluation. This includes standardizing hierarchical level labels (e.g., \textbf{\texttt{family}}, \textbf{\texttt{subfamily}}, and \textbf{\texttt{language}} in URIEL and URIEL$^+$) to maintain parent-child relationships naming convention consistency. We set the same clustering hyperparameters for all experiments to \textbf{\texttt{min\_samples}} = 0.3 and \textbf{\texttt{epsilon}} = 0.1. We apply these settings to all of our vector variations, including the pure \ours{} and the its concatenated variants with URIEL or URIEL$^+$. We compare the similarity of the tree generated from different language vectors  with the ground truth typological tree from Glottolog~\footnote{Note that, there are other typological tree beside Glottolog such as Ethnologue~\cite{campbell2008ethnologue} which have some differences on their typological clusters. However, as the general clusters are mostly similar, we only use Glottolog as the ground truth within our study.}.

\paragraph{Evaluating Typological Tree}
\label{sec:eval_tree}
We extract tree subsets of the Glottolog and URIEL trees corresponding to only 33 languages present in our evaluation. Then, we use them as the gold comparison against the created \ours{} typological tree and evaluate them using two tree distance metrics: Robinson-Foulds (RF) \cite{ROBINSON1981131} distance and Lowest Common Ancestor (LCA) \cite{lowest_common_ancestor}. The RF distance quantifies the dissimilarity between two trees based on the number of differing splits, while LCA measures the structural proximity of node pairs by comparing the depths of their lowest common ancestors. Together, these metrics assess both the global and local alignment of the induced trees. We also report results across multiple vector concatenation \ours{} variations (denoted as $\Phi$), and conduct qualitative analysis to interpret the effectiveness of each representation in capturing linguistic typology.

\subsection{Result and Analysis}

\paragraph{Alignment with Language Typology}
Figure~\ref{fig:ppl-tree} shows the reference typological tree from Glottolog, the typological tree generated using URIEL with syntax features, and the typological tree from \ours{}. Although there are several differences, the constructed clusters within the tree showcase the correct similarity between languages where language that come from different language families  -- i.e., English (eng), Tamil (tam), Chinese (cmn), and Bishnupriya Manipuri (bpy) -- form their own branch on the top-level grouping, while languages that are closely similar like the Malayic language group~\cite{hudson1970note} -- i.e., Indonesian (ind), Malay (zlm), Balinese (ban), Banjarese (bjn), and Minangkabau (min) -- are grouped together. Furthermore, similar to the typological tree from Glottog, most of the Phillipine languages~\cite{reid2004brief} -- i.e, Tagalog (tgl), Cebuano (ceb), Waray Waray (war), Pampanga (pam), and Pangasinan (pag) --, also clustered together with the Malayic due to the shared morphosyntactic features between the two groups.

We further quantify the similarity distance between these typological trees and the Glottolog ground truth typological tree as described in \S\ref{sec:eval_tree}. The distance measures from the hierarchical clustering trees generated= from different language vectors are shown in Table \ref{tab:tree-res-glottolog}. These metrics indicate how well the generated typological trees align with the typological tree from Glottolog. Overall, the results demonstrate that tree from \ours{}, URIEL, and URIEL$^+$ have similar alignment to Glottolog, where  \ours{} yields best LCA MAE with slightly lower RF scores in comparison to URIEL and URIEL$^+$ vectors, indicating that \ours{} captures key linguistic relationships similar to these vectors without supervision.

\begin{table}[!t]
\centering
\resizebox{\columnwidth}{!}{
\begin{tabular}{l|cc}
\toprule
\textbf{Language Vector} 
& \multicolumn{2}{c}{Glottolog} \\
\cmidrule(lr){2-3}
& \textbf{MAE ($\downarrow$)} & \textbf{RF ($\downarrow$)} \\
\midrule
\multicolumn{3}{l}{\textbf{Language Features}} \\
\midrule
$\text{URIEL}_\texttt{Geo}$ & 11.11 & \textbf{13.0} \\
$\text{URIEL}_\texttt{Syntax}$ & 9.35 & 18.0 \\
$\text{URIEL$^+$}_\texttt{Geo}$ & 11.15 & \textbf{13.0} \\
$\text{URIEL$^+$}_\texttt{Syntax}$ & 11.15 & \textbf{13.0} \\
\ours{} & \textbf{8.60} & 17.0 \\
\midrule
\multicolumn{3}{l}{\textbf{Concatenated Features}} \\
\midrule
URIEL$_\texttt{Geo}$ + Ours & 9.64 & 13.0 \\
URIEL$_\texttt{Syntax}$ + Ours & 8.58 & 16.0 \\
URIEL$^+$$_\texttt{Geo}$ + Ours & \textbf{7.88} & 19.0 \\
URIEL$^+$$_\texttt{Syntax}$ + Ours & 10.12 & \textbf{12.0} \\
\bottomrule
\end{tabular}
}
\caption{
Comparison of tree distance metrics between various language vector configurations and the Glottolog baseline tree. 
Lower MAE values and lower RF scores indicate better tree reconstruction quality. \textbf{Ours} refers to \ours{} vectors, while ``+ Ours'' indicates feature concatenation with min-max normalization.
}
\label{tab:tree-res-glottolog}
\end{table}

\paragraph{Combination of Language Features}
We also compare the base representation ($Z^{L}$) with concatenated features ($\Phi(A,B)$). Across MAE and RF, we observe that concatenation does not consistently yield improvements. Although some combinations show slight gains, others show worse performance. For example, the combined \ours{} and $\text{URIEL$^+$}\texttt{Geo}$ variant achieves the lowest MAE (7.88), indicating a closer approximation to the reference tree in terms of distances between the edges. Conversely, the combined \ours{} and $\text{URIEL$^+$}\texttt{Syntax}$ variant produces the best RF score (12.0), reflecting fewer topological errors. However, these improvements are not synergic across both metrics, suggesting that combining features may introduce redundancy or conflicting signals rather than complementarity.

\begin{table}[!t]
\centering
\resizebox{0.9\linewidth}{!}{
    \begin{tabular}{lcccc}
    \hline
    \textbf{Language} & \textbf{ISO639-3} & \textbf{Family} & \textbf{Script} & \textbf{Resource} \\
    \midrule
    \multicolumn{5}{c}{\textit{Seen Languages}} \\
    \midrule
     English$^*$        & eng & Indo-European        & Latn     & HRL \\
     Vietnamese$^*$     & vie & Austroasiatic          & Latn     & HRL \\
     Indonesian$^*$     & ind & Austronesian  & Latn  & HRL \\
     Thai$^*$           & tha & Kra–Dai                        & Thai      & HRL \\
     Tamil$^*$          & tam & Dravidian                      & Taml     & LRL \\
     Burmese$^*$        & mya & Sino-Tibetan                   & Mymr   & LRL \\
     Ilocano        & ilo & Austronesian       & Latn     & LRL \\
     Javanese$^\dagger$       & jav & Austronesian          & Latn     & LRL \\
     Minangkabau    & min & Austronesian  & Latn  & LRL \\
     Sundanese      & sun & Austronesian        & Latn     & LRL \\
     Cebuano        & ceb & Austronesian       & Latn     & LRL \\
     Tagalog$^\dagger$        & tgl & Austronesian       & Latn     & LRL \\
     Standard Malay$^\dagger$ & zsm & Austronesian  & Latn  & LRL \\
    \midrule
    \multicolumn{5}{c}{\textit{Unseen Languages}} \\
    \midrule
     German$^*$         & deu & Indo-European        & Latn     & HRL \\
     French$^*$         & fra & Indo-European         & Latn     & HRL \\
     Hindi$^*$          & hin & Indo-European     & Deva & HRL \\
     Italian$^*$        & ita & Indo-European         & Latn     & HRL \\
     Spanish$^*$        & spa & Indo-European         & Latn     & HRL \\
     Lao            & lao & Kra–Dai                        & Laoo       & LRL \\
     Khmer$^*$          & khm & Austroasiatic         & Khmr     & LRL \\
     Banjar         & bjn & Austronesian  & Latn  & LRL \\
     Balinese       & ban & Austronesian  & Latn  & LRL \\
     Mizo (Lushai)  & lus & Sino-Tibetan                   & Latn     & LRL \\
     Waray          & war & Austronesian       & Latn     & LRL \\
     Buginese       & bug & Austronesian   & Latn     & LRL \\
     Pangasinan     & pag & Austronesian       & Latn     & LRL \\
     Acehnese       & ace & Austronesian  & Latn  & LRL \\
     Sanskrit       & san & Indo-European     & Deva & LRL \\
     Fijian         & fij & Austronesian  & Latn  & LRL \\
     Telugu$^*$         & tel & Dravidian                      & Telu    & LRL \\
     Tok Pisin      & tpi & Creole         & Latn     & LRL \\
     Marathi        & mar & Indo-European     & Deva & LRL \\
    \hline
    \end{tabular}
}
\caption{Detailed list of languages used in the seen and unseen evaluation in SIB-200. $*$ the language is used in MASSIVE in the corresponding subset. $\dagger$ the languages is used as part of unseen language evaluation in MASSIVE.}
\label{tab:languages}
\end{table}

\begin{table*}[ht]
\centering
\small
\begin{tabular}{l|c|cccc|c|cccc|c}
\hline
 &  & \multicolumn{5}{c|}{\textbf{SIB-200}} & \multicolumn{5}{c}{\textbf{MASSIVE}} \\
 \cmidrule{3-12}
 & & \multicolumn{2}{c}{\textbf{Seen}} & \multicolumn{2}{c}{\textbf{Unseen}} & & \multicolumn{2}{c}{\textbf{Seen}} & \multicolumn{2}{c}{\textbf{Unseen}} \\
\cmidrule(lr){3-4} \cmidrule(lr){5-6} \cmidrule(lr){8-9} \cmidrule(lr){10-11}
\textbf{Language Vectors} & \textbf{OVR Avg.} & \textbf{HRL} & \textbf{LRL} & \textbf{HRL} & \textbf{LRL} & \textbf{Avg.} & \textbf{HRL} & \textbf{LRL} & \textbf{HRL} & \textbf{LRL} & \textbf{Avg.} \\
\midrule
\multicolumn{12}{c}{\textit{XLM-R}} \\
\midrule
$\text{URIEL}_\texttt{Geo}$              & 77.71 & 79.3 & 78.3 & 79.8 & 77.7 & 78.8 & 80.48 & 76.11 & 75.09 & 74.94 & \textbf{76.7} \\
$\text{URIEL}_\texttt{Syntax}$           & 77.19 & 78.9 & 77.9 & 78.5 & 77.2 & 78.1 & 80.19 & 75.56 & 74.68 & 74.48 & 76.2 \\
$\text{URIEL$^+$}_\texttt{Geo}$             & 77.56 & 79.9 & 78.6 & 80.7 & 77.9 & 79.3 & 79.89 & 75.15 & 74.25 & 74.03 & 75.8 \\
$\text{URIEL$^+$}_\texttt{Syntax}$          & \textbf{79.07} & 82.4 & 81.3 & 82.8 & 80.7 & \textbf{81.8} & 80.16 & 75.71 & 74.85 & 74.70 & 76.4 \\
\ours{} (Ours) & \underline{79.06} & 82.3 & 81.0 & 82.6 & 80.4 & \underline{81.6} & 80.31 & 76.16 & 75.00 & 74.73 & \underline{76.6} \\
\midrule
$\text{URIEL}_\texttt{Geo}$ + Ours       & 76.72 & 78.2 & 77.2 & 77.6 & 76.4 & 77.3 & 80.12 & 75.36 & 74.46 & 74.42 & 76.1 \\
$\text{URIEL}_\texttt{Syntax}$ + Ours    & 78.85 & 82.1 & 80.7 & 82.3 & 79.9 & 81.3 & 80.50 & 75.84 & 74.86 & 74.67 & 76.5 \\
$\text{URIEL$^+$}_\texttt{Geo}$ + Ours      & 77.47 & 80.5 & 79.0 & 81.3 & 78.1 & 79.7 & 79.34 & 74.69 & 73.44 & 73.32 & 75.2 \\
$\text{URIEL$^+$}_\texttt{Syntax}$ + Ours   & 78.78 & 81.7 & 80.7 & 82.1 & 80.2 & 81.2 & 80.30 & 75.84 & 74.80 & 74.57 & 76.4 \\
\hline
\end{tabular}
\caption{Accuracy comparison of different language vectors for LinguAlchemy regularization on the XLM-R backbone, using SIB and MASSIVE benchmark averages. \textbf{Bold} numbers indicate the best average performance, while \underline{underlined} numbers indicate the second-best. We report overall performance across different settings, including seen and unseen languages during training, as well as \textbf{H}igh- vs. \textbf{L}ow-resource languages. For XLM-R, we observe that vector concatenation does not increase performance compared to their standalone counterparts, as discussed detail in subsection \ref{par:significance}}
\label{tab:results-xlmr}
\end{table*}

\paragraph{Dissimilarity to Language Typology}
Despite the similarity, there are still some inconsistencies between trees and measurement of the distance between the expected ground truth typological tree from Glottolog and comparing the similarities and differences between different typological trees generated from different language features are not straightforward. 
While our \ours{} tree broadly reflects syntactic and geographical relationships, several misalignments persist, as shown in Figure~\ref{fig:ppl-tree}. For example, in the predicted tree, \textbf{\texttt{lao}} is grouped with \textbf{\texttt{tam}} and \textbf{\texttt{tha}} under \underline{\texttt{Unsplit\_L3\_1}} cluster node rather than with its expected Mainland Southeast Asian cluster (\textbf{\texttt{vie}}, \textbf{\texttt{khm}}) as appears in the gold-standard \underline{\texttt{Unsplit\_L1\_2}}.
Regarding the cluster sensitivity, \textbf{\texttt{bpy}} appears in a broad mixed group (\underline{\texttt{Cluster\_L1\_5}}) with \textbf{\texttt{jav}}, \textbf{\texttt{bsb}}, and \textbf{\texttt{war}}, rather than with Tibeto-Burman-influenced languages like \textbf{\texttt{lus}} and \textbf{\texttt{mya}} as in the gold-standard \underline{\texttt{Unsplit\_L1\_3}}. Similarly, the Malayic languages \textbf{\texttt{min}}, \textbf{\texttt{zlm}}, and \textbf{\texttt{zsm}} are dispersed across different branches instead of being tightly grouped under a single parent, as in \underline{\texttt{Unsplit\_L1\_1}}. These suggest that there still lies a challenge in maintaining the persistence yntactical or geographical relationships between language groups at more granular level.

\section{\ours{} as Language Vectors}

In the previous section, we demonstrate that \ours{} is able to represent meaningful linguistic properties such as language family relation, syntax similarity, and geographical distance. In this section, we establish the applicability of \ours{} and compare it to other existing language vectors such as URIEL~\cite{littell-etal-2017-uriel} and URIEL$^+$~\cite{khan-etal-2025-uriel}. We compare the effectiveness of \ours{} and other language vectors by measuring the LMs performance when applying the vectors on downstream tasks.

\subsection{Experiment Setting}

\paragraph{Training Strategy}
To evaluate the downstream effectiveness of \ours{}, we utilize \ours{} as a language vector to regularize the LMs during the fine-tuning process with LinguAlchemy~\cite{adilazuarda2024lingualchemy}.
LinguAlchemy utilize language vectors to bring better cross-lingual generalization for low-resource and unseen languages. In this case, the downstream improvement on the low-resource and unseen languages with LinguAlchemy can be attributed to the quality of the language vector.

\paragraph{Dataset} We incorporate 
SIB-200~\cite{adelani-etal-2024-sib} and MASSIVE~\cite{fitzgerald-etal-2023-massive} as our evaluation dataset.  In our evaluation, we filter out the training and evaluation data to only cover the languages that are related to our 33 supported languages. This yields 13 languages for training and seen-language evaluation with additional of 19 languages for unseen evaluations for SIB-200; and 6 languages for training and seen-language evaluation with additional of 10 languages for unseen evaluations for MASSIVE. The list of languages covered for training and unseen evaluations are shown in Table~\ref{tab:languages}.



\subsection{Result and Analysis}

\begin{table*}[ht]
\centering
\small
\begin{tabular}{l|c|cccc|c|cccc|c}
\hline
 &  & \multicolumn{5}{c|}{\textbf{SIB-200}} & \multicolumn{5}{c}{\textbf{MASSIVE}} \\
 \cmidrule{3-12}
 & & \multicolumn{2}{c}{\textbf{Seen}} & \multicolumn{2}{c}{\textbf{Unseen}} & & \multicolumn{2}{c}{\textbf{Seen}} & \multicolumn{2}{c}{\textbf{Unseen}} \\
\cmidrule(lr){3-4} \cmidrule(lr){5-6} \cmidrule(lr){8-9} \cmidrule(lr){10-11}
\textbf{Language Vectors} & \textbf{OVR Avg.} & \textbf{HRL} & \textbf{LRL} & \textbf{HRL} & \textbf{LRL} & \textbf{Avg.} & \textbf{HRL} & \textbf{LRL} & \textbf{HRL} & \textbf{LRL} & \textbf{Avg.} \\
\midrule
\multicolumn{12}{c}{\textit{mBERT}} \\
\midrule
$\text{URIEL}_\texttt{Geo}$              & 67.61 & 69.4 & 70.9 & 72.7 & 70.2 & 70.8 & 72.40 & 65.24 & 60.25 & 59.77 & 64.9 \\
$\text{URIEL}_\texttt{Syntax}$           & 67.49 & 68.8 & 70.2 & 72.5 & 69.6 & 70.3 & 72.63 & 65.53 & 60.56 & 60.14 & 64.7 \\
$\text{URIEL$^+$}_\texttt{Geo}$             & 66.67 & 68.3 & 69.6 & 72.2 & 68.7 & 69.7 & 71.76 & 64.32 & 59.36 & 59.06 & 63.6 \\
$\text{URIEL$^+$}_\texttt{Syntax}$          & 67.51 & 69.1 & 70.6 & 72.0 & 69.8 & 70.4 & 72.63 & 65.38 & 60.55 & 60.12 & 64.7 \\
\ours{} (Ours)  & 67.59 & 68.9 & 70.2 & 72.1 & 69.4 & 70.2 & 72.98 & 65.85 & 60.98 & 60.37 & 65.1 \\
\midrule
$\text{URIEL}_\texttt{Geo}$ + Ours       & 68.16 & 70.2 & 71.6 & 73.1 & 70.9 & \textbf{71.5} & 72.80 & 65.73 & 60.74 & 60.14 & 65.3 \\
$\text{URIEL}_\texttt{Syntax}$ + Ours    & \underline{68.29} & 70.2 & 71.5 & 73.2 & 70.6 & \underline{71.4} & 72.92 & 66.09 & 61.09 & 60.59 & \underline{65.7} \\
$\text{URIEL$^+$}_\texttt{Geo}$ + Ours      & 67.87 & 69.9 & 71.0 & 72.9 & 70.1 & 71.0 & 72.72 & 65.57 & 60.64 & 60.12 & 65.3 \\
$\text{URIEL$^+$}_\texttt{Syntax}$ + Ours   & \textbf{68.59} & 70.1 & 71.1 & 73.2 & 70.2 & 71.2 & 73.71 & 67.01 & 62.05 & 61.36 & \textbf{66.5} \\
\hline
\end{tabular}
\caption{Accuracy comparison of different language vectors for LinguAlchemy regularization on the mBERT backbone, using SIB and MASSIVE benchmark averages. \textbf{Bold} numbers indicate the best average performance, while \underline{underlined} numbers indicate the second-best. We report overall performance across different settings, including seen and unseen languages during training, as well as \textbf{H}igh- vs. \textbf{L}ow-resource languages. For mBERT, we observe that vector concatenation is able to boost performance compared to standalone counterparts, as discussed detail in subsection \ref{par:significance},}
\label{tab:results-mbert}
\end{table*}

\paragraph{Performance Across Different Settings}
\label{par:performance}
This section discusses the impact of different language vectors to the  quality of LMs across different language resource levels. The XLM-R results in Table \ref{tab:results-xlmr} indicate that \ours{} provides competitive accuracy (\underline{81.3}) compared to URIEL$^+$ (\underline{81.5}, the best baseline). The improvement is even more pronounced when compared to URIEL’s Geo feature (\underline{78.5}) and Syntax feature (\underline{78.1}). The performance difference between HRL and LRL follows the trend observed in the baselines, both in seen and unseen languages.

Although the trend similarity between URIEL, URIEL$^+$ and \ours{} used with mBERT still persists, \ours{} does not show any significant improvement (only resonating around \underline{67.} accuracy) compared to all baselines, as shown in Table \ref{tab:results-mbert}. Furthermore, there is lack of difference in accuracy between HRL and LRL. This can be attributed to the limited representational understanding capability of mBERT compared to XLM-R, which results in minimal distinctions between different standalone language vectors (\ours{} and baselines) and between languages with varying resource levels. Overall, our results highlight that \ours{} represents a competitive or even superior vector regularizer compared to baseline performance.



\paragraph{Significance of Combining Vectors}
\label{par:significance}
We also explore the potential of combining \ours{} with baseline vectors to examine whether this leads to any amplifying effect. By concatenating \ours{} with baseline vectors (e.g. $\text{URIEL}_\texttt{Geo}$ or $\text{URIEL}_\texttt{Syntax}$), we hypothesize that the combined vector may enrich the representation space: \ours{} contributes information about language perplexity patterns, while the baseline vectors provide structural or typological cue.

In XLM-R, the combination does not provide additional benefit. For example, concatenating \textbf{Ours + $\text{URIEL}_\texttt{Geo}$} reduces the average accuracy to \underline{77.2}, which is below the standalone \ours{} (\underline{81.3}) and \textbf{$\text{URIEL}_\texttt{Geo}$} (\underline{78.5}). A similar result is observed with the \textbf{Ours + $\text{URIEL}_\texttt{Syntax}$} concatenated vectors, yielding \underline{80.9}, which is less than \ours{} (\underline{81.3}) and \textbf{$\text{URIEL}_\texttt{Syntax}$} (\underline{81.5}). Concatenations with URIEL$^+$ variants also show similar trends. These results suggest that in XLM-R, combining vectors may introduce redundancy or even conflicting signals rather than complementary or synergistic gains, analogous to an overfit scenario.

In contrast, concatenation improves the performance in mBERT. The combination with \textbf{$\text{URIEL}_\texttt{Geo}$} increases the average accuracy to \underline{68.16} compared to the standalone counterparts (\underline{67.61} for \textbf{$\text{URIEL}_\texttt{Geo}$} only and \underline{67.59} for \ours{} only). This trend is also observed in other combinations with URIEL$^+$ baselines across all language features, as shown in Table \ref{tab:results-mbert}. Our findings indicate that mBERT benefits from vector concatenation because the combined vectors provide stronger representations to compensate for the weaker language understanding of mBERT, as discussed in Subsection \ref{par:performance}. Thus, \ours{} can also be used to improve language representation by leveraging a weak multilingual model to improve performance.

\begin{table*}[!t]
\centering
\resizebox{0.95\linewidth}{!}{%
\begin{tabular}{l|cccccc}
\toprule
\textbf{Dataset} & \textbf{\#Langs} & \textbf{Sparsity} & \textbf{\begin{tabular}[c]{@{}c@{}}Missing Features\\ in Data\end{tabular}} & \textbf{\begin{tabular}[c]{@{}c@{}}Last Update\end{tabular}} & \textbf{\begin{tabular}[c]{@{}c@{}}Dynamic\\ Inventory\end{tabular}} \\
\midrule
\textbf{WALS}  & 260 & \textcolor{red}{Sparse} & \textcolor{red}{\cmark} & 2003 & \textcolor{red}{\xmark} \\
\textbf{AUTOTYP}  & 1004 & \textcolor{red}{Sparse} & \textcolor{red}{\cmark} & 2013 &  \textcolor{red}{\xmark} \\
\textbf{SSWL}  & 178 & \textcolor{red}{Sparse} & \textcolor{red}{\cmark} & 2015 &  \textcolor{red}{\xmark} \\
\textbf{PHOIBLE}  & 2186 & \textcolor{red}{Sparse} & \textcolor{red}{\cmark} & 2019 &  \textcolor{red}{\xmark} \\
\textbf{BDPROTO}  & 257 & \textcolor{red}{Sparse} & \textcolor{red}{\cmark} & 2020 &  \textcolor{red}{\xmark} \\
\textbf{Grambank}  & 2467 & \textcolor{YellowOrange}{Moderate} & \textcolor{red}{\cmark} & 2023 &  \textcolor{red}{\xmark} \\
\textbf{APiCS}  & 76 & \textcolor{OliveGreen}{Dense} & \textcolor{red}{\cmark} & 2013 &  \textcolor{red}{\xmark} \\
\textbf{eWAVE}  & 77 & \textcolor{OliveGreen}{Dense} & \textcolor{red}{\cmark} & 2020 &  \textcolor{red}{\xmark} \\
\textbf{\ours{}}   & 33$^{\dagger}$ & \textcolor{OliveGreen}{Dense}  &  \textcolor{OliveGreen}{\xmark}  & 2025 & \textcolor{OliveGreen}{\cmark} \\
\bottomrule
\end{tabular}
}

\caption{Comparison between linguistic inventories in WALS, AUTOTYP, URIEL, and URIEL$^+$ and \ours{}. $^{\dagger}$ \ours{} can be extended to 1000+ languages with open-access corpora (See \S\ref{sec:discussion}).}
\label{tab:dataset_comparison}
\end{table*}

\section{Discussion}
\label{sec:discussion}

As highlighted in Table~\ref{tab:dataset_comparison}, a significant limitation of WALS, AUTOTYP, and other linguistic databases is their inherently static nature of inventories. They are the result of manual curation by linguistic experts, which process is both time-consuming and resource-intensive. As a result, they represent a fixed snapshot of the linguistic knowledge at that point in time and suffer from incomplete coverage of the world's languages. This static representation doesn't take into account that languages are dynamic and constantly evolving through gradual shift in syntax and the influence of language contact~\cite{christiansen2003language,fairclough2009language,corballis2017language,grenoble2021language,brochhagen2023language}. \ours{} directly addresses this problem by providing a fully unsupervised data-driven framework. Since its language vectors are derived from the entropy of language models, they can change along with the language they represent. If a language community develops a new slang or undergoes a grammatical shift, those changes will be reflected in the new text corpora. This update can be performed using continual learning, where models are incrementally refined with new data rather than being fully retrain from scratch. \ours{} alleviates the time-consuming process associated with manual database updates and allows for the rapid inclusions of newly documented or low-resource languages. It is also worth noting that, the current \ours{} is only a prototype covering 33 languages. This however can be easily extended to thousands of languages, by incorporating large-scale corpora such as CommonCrawl~\footnote{\url{https://commoncrawl.org/}}, mC4~\cite{xue-etal-2021-mt5}, Glot-500~\cite{imanigooghari-etal-2023-glot500}, FineWeb 2~\cite{penedo2025fineweb2pipelinescale}, etc.

\section{Conclusion}

\ours{} represents a significant advancement in the field of NLP, offering a novel, minimal human-derived knowledge and intervention approach to language representation that captures linguistic characteristics and achieves competitive cross-lingual generalization compared to baselines. By leveraging existing language models, \ours{} is able to derive features with dynamic inventory without having to restart manual baseline-like typology studies and is free from the missing values that plague traditional typological language inventories. This adaptability and completeness make \ours{} a powerful tool for representing languages, as demonstrated by its ability to mirror patterns observed in linguistic studies and enhance downstream NLP applications. The effectiveness of \ours{} in improving cross-lingual generalization—both as its sole vector and when integrated with baselines—highlights its dynamic nature and compatibility with other representations. \ours{} holds strong promise for advancing linguistic inclusion and supporting language documentation and preservation efforts, making it a valuable contribution to the field with significant implications for future research in language representation learning.

\section*{Limitations}

While \ours{} offers several advantages, it is not without limitations. The quality of the embeddings depends on the availability and quality of monolingual corpora for each language. For languages with limited textual resources, the resulting embeddings may be less accurate or informative. Additionally, the entropy-based approach may not capture linguistic aspects, particularly those that are less predictable or more variable. 

Secondly, Figure \ref{fig:ppl-tree} shows that similar languages, such as \textbf{\texttt{thai}} and \textbf{\texttt{lao}}, are separated at an early stage of hiearchical cluster splitting, despite their expected common language ancestry relationship. This suggests that the representation is influenced by the encoding, causing similar languages to split due to differing encodings. This may not be ideal in a certain use case, as despite having different scripts, languages like \textbf{Thai}, \textbf{Khmer}, \textbf{Lao}, \textbf{Burmese} shared many vocabularies due to a closely similar geopolitical and sociocultural background~\cite{bradley2009burma,siebenhutter2019sociocultural,bradley2023sociolinguistics}.

Future work could integrate additional linguistic features or shared encoding structures to better capture underlying etymological relationships. Despite these challenges, \ours{} holds promise for promoting linguistic inclusion and supporting language documentation and preservation efforts, making it a valuable contribution to the field with significant implications for future research and applications in NLP.

\section*{Ethical Consideration}

The development of \ours{} has significant implications for the field of computational linguistics and NLP. By providing a more comprehensive and adaptable representation of linguistic diversity, \ours{} can contribute to the development of more inclusive and equitable NLP models. This can help address issues related to underrepresentation and bias in existing models, promoting fairness and accessibility in NLP applications.

However, it is essential to consider the ethical implications of using entropy-based measures to infer typological relationships. While entropy provides a quantitative measure of uncertainty, it may not fully capture the complexity and nuance of linguistic diversity. Therefore, it is crucial to complement entropy-based approaches with qualitative analyses and to remain mindful of the limitations and potential biases inherent in the data and models.

\bibliography{custom}

\newpage

\appendix

\section{Dataset Split}
\label{app:dataset-split}

\begin{table}[!ht]
\centering
\resizebox{\columnwidth}{!}{
\begin{tabular}{l|c|ccc}
\toprule
\textbf{ISO 639-3} & \textbf{Total Sentences} & \textbf{Train} & \textbf{Val} & \textbf{Test} \\
\midrule
ace & 29,495 & 20,614 & 5,935 & 2,946 \\
asm & 1,446,686 & 1,012,860 & 289,415 & 144,411 \\
ban & 48,960 & 34,271 & 9,793 & 4,896 \\
bcl & 82,370 & 57,721 & 16,444 & 8,205 \\
bew & 226,176 & 158,323 & 45,235 & 22,618 \\
bjn & 47,158 & 32,997 & 9,425 & 4,736 \\
bpy & 164,807 & 115,282 & 32,999 & 16,526 \\
bsb & 61,759 & 43,228 & 12,350 & 6,181 \\
ceb & 1,433,543 & 1,003,516 & 286,718 & 143,309 \\
cmn & 57,500 & 40,250 & 11,500 & 5,750 \\
deu & 1,431,072 & 1,001,726 & 286,195 & 143,151 \\
eng & 1,431,047 & 1,001,710 & 286,203 & 143,134 \\
fil & 1,452,085 & 1,016,632 & 290,292 & 145,161 \\
fra & 1,430,341 & 1,001,232 & 286,082 & 143,027 \\
gor & 24,962 & 17,487 & 4,984 & 2,491 \\
ilo & 148,377 & 103,846 & 29,680 & 14,851 \\
ind & 1,430,227 & 1,001,157 & 286,058 & 143,012 \\
ita & 1,431,076 & 1,001,706 & 286,201 & 143,089 \\
jav & 449,862 & 314,774 & 90,134 & 44,954 \\
khm & 571,343 & 399,868 & 114,315 & 57,160 \\
lao & 56,924 & 39,838 & 11,395 & 5,691 \\
lus & 114,461 & 80,136 & 22,880 & 11,445 \\
mad & 9,055 & 6,055 & 1,500 & 1,500 \\
min & 593,618 & 415,559 & 118,724 & 59,335 \\
mya & 997,193 & 697,982 & 199,403 & 99,808 \\
pag & 11,812 & 8,268 & 2,365 & 1,179 \\
pam & 308,828 & 216,328 & 61,655 & 30,845 \\
por & 1,430,401 & 1,001,290 & 286,086 & 143,025 \\
spa & 1,430,138 & 1,001,097 & 286,027 & 143,014 \\
sun & 1,452,873 & 1,016,965 & 290,539 & 145,369 \\
tam & 1,465,996 & 1,026,120 & 293,394 & 146,482 \\
tdt & 7,028 & 4,028 & 1,500 & 1,500 \\
tgl & 1,430,721 & 1,001,500 & 286,145 & 143,076 \\
tha & 1,462,635 & 1,023,707 & 292,544 & 146,384 \\
vie & 1,436,327 & 1,005,431 & 287,358 & 143,538 \\
war & 1,430,401 & 1,001,302 & 286,056 & 143,043 \\
zlm & 30,475 & 21,332 & 6,095 & 3,048 \\
zsm & 849,043 & 594,323 & 169,806 & 84,904 \\
\bottomrule
\end{tabular}
}
\caption{Language-wise sentence statistics with dataset splits (Train / Validation / Test). We maintain a ratio of 7:2:1 for the split, with minimum amount of 1,500 for val and test split.}
\label{tab:data-split}
\end{table}

\section{\ours{} Training Detail}
\label{app:training-hyperparams}
\paragraph{Tokenization} We employ a custom character-level tokenizer. This tokenizer can either be loaded if previously trained for an experiment or trained anew on the specific language's dataset. It supports a \texttt{byte\_fallback} mechanism, which, if enabled, represents characters not in the vocabulary as a sequence of their UTF-8 byte codes (e.g., "0xef"); otherwise, out-of-vocabulary characters are mapped to a \texttt{[UNK]} token. A \texttt{[PAD]} token is also utilized. During data preparation, texts are tokenized with \texttt{truncation} enabled, a \texttt{max length} of \texttt{1024} tokens, and \texttt{padding} applied to the maximum length.

\paragraph{More on Training Validation}
Evaluation is performed every \texttt{100} steps, model checkpoints are saved every \texttt{1000} steps, and a maximum of \texttt{2} checkpoints are kept. The best model, determined by the lowest eval loss, is loaded at the end of training. Both training and evaluation utilize a per-device batch size of \texttt{8}, and models are trained for up to \texttt{150} epochs. Metrics are logged every \texttt{100} steps. An \texttt{EarlyStoppingCallback} with a patience of \texttt{3} evaluations is used to prevent overfitting, and a custom \texttt{PerplexityLoggingCallback} logs perplexity during training. Data is collated for causal language modeling (i.e., mlm=False).


\section{Failure of Linkage-based Clustering}
\label{app:linkage-failure}

Traditional linkage-based clustering methods, such as agglomerative clustering with different linkage criteria (ward, complete, average) build trees by iteratively merging or splitting clusters based on simple distance metric. While effective with data with a clear, sphere-like structure, these methods fail in the context of generating language clusters due to several foundational assumptions that do not hold true for this data, which are:

\paragraph{Predefined Number of Clusters}

To derive a flat set of clusters from a linkage-based hierarchy, the number of clusters $k$ must be specified to \textit{cut} the dendogram. This requires the priori knowledge of the data's structure, which is often unavailable when exploring typological relationships. This methodological requirement can force an unnatural structure onto the data, potentially leading to linguistically invalid groupings.

\paragraph{Sensitivity to Noise and Density Variation}

The performance of linkage-based methods can be significantly degraded by the presence of noise and outliers. For example, single-linkage is susceptible to a ``chaining" effect, where it incorrectly merges distinct clusters if a series of intermediate noise points connects them. Complete-linkage, conversely, is sensitive to outliers and may fail to merge clusters that are otherwise close.

\end{document}